\pdfoutput=1
\documentclass[11pt]{article}

\usepackage{acl}

\usepackage{times}
\usepackage{latexsym}
\usepackage{amsmath,amssymb,amsfonts}
\usepackage{algorithmic}
\usepackage{booktabs}
\usepackage{graphicx}
\usepackage{float}
\usepackage[font=scriptsize,labelfont=bf]{caption}
\usepackage[font=scriptsize,labelfont=bf]{subcaption}
\usepackage{textcomp}
\usepackage{xcolor}
\usepackage{colortbl}
\usepackage[utf8]{inputenc}
\usepackage[all]{nowidow}

\usepackage[T1]{fontenc}
\usepackage[utf8]{inputenc}

\usepackage{microtype}

\makeatletter
\newcommand\nocaption{%
    \renewcommand\p@subfigure{}
    \renewcommand\thesubfigure{\thefigure\alph{subfigure}}
}
\makeatother

\title{Compositional Generalization in
Grounded Language Learning via Induced Model Sparsity}

\author{{Sam Spilsbury} \and Alexander Ilin \\
\textit{Department of Computer Science} \\
\textit{Aalto University}\\
Espoo, Finland \\
\texttt{\{first.last\}@aalto.fi}
}

\begin{document}
\maketitle
\begin{abstract}
We provide a study of how induced model sparsity can help achieve compositional generalization and better sample efficiency in grounded language learning problems. We consider simple language-conditioned navigation problems in a grid world environment with disentangled observations. We show that standard neural architectures do not always yield compositional generalization. To address this, we design an agent that contains a goal identification module that encourages sparse correlations between words in the instruction and attributes of objects, composing them together to find the goal.\footnote{\href{https://github.com/aalto-ai/sparse-compgen}{github.com/aalto-ai/sparse-compgen}} The output of the goal identification module is the input to a value iteration network planner. Our agent maintains a high level of performance on goals containing novel combinations of properties even when learning from a handful of demonstrations. We examine the internal representations of our agent and find the correct correspondences between words in its dictionary and attributes in the environment.
\end{abstract}

\section{Introduction}
\label{sec:introduction}
Ideally, when training an agent that acts upon natural language 
instructions, we want the agent to understand the meaning
of the words, rather than overfitting to the training instructions.
We expect that when an agent encounters an unfamiliar
instruction made up of familiar terms, it should be able to complete
the task. In this sense, the agent learns to leverage both \textit{groundedness} of language;
for example in English, tokens in the language map to observed
attributes of objects or phenomena in its environment, as well
as its \textit{compositionality}; which enables the description
of potentially infinite numbers of
new phenomena from known components \cite{book/chomsky/65}. Using
language to express goals is potentially a way to approach
task distribution shift and sample efficiency, key problems in
reinforcement learning \cite{conf/icml/Sodhani0P21, conf/corl/JangIKKELLF21}.

However, compositional generalization does not come automatically with standard architectures when
using language combined with multi-modal inputs, as indicated by the mixed results of
generalization performance in \citet{journals/corr/abs-2106-02972/Goyal/2021, conf/icml/Sodhani0P21}.
Concurrently with \citet{conf/emnlp/QiuH0SS21}, we show
that the Transformer architecture can demonstrate
generalization, but requires large amounts of data for training.
In this work, we tackle sample inefficiency and retain generalization.

Our contributions are as follows. We propose a model and a training method that utilizes the inductive biases of \textit{sparse interactions}
and \textit{factor compositionality} when finding relationships between words and disentangled attributes. We hypothesize that such sparsity in the \textit{interactions}
between object attributes and words (as opposed to just their representations) leads to
a correct identification of what attributes the words actually correspond to, instead
of what they are merely correlated with. We show in both quantitative and qualitative
experiments that such sparsity and factor compositionality enable compositional
generalization. To improve sample efficiency, we decouple the goal identification task
(which requires language understanding) from the planning process (implemented with an 
extension of Value Iteration Networks).

\section{Related Work}
\label{sec:related}

% one sentence motivation of why this is important in GLL
\paragraph{Compositional Generalization and Language Grounding} 
There is a long line of work on learning to achieve language
encoded instructions within interactive environments.
Vision-Language Navigation environments
typically require an agent to navigate to a requested
goal object (for example, DeepMind Lab
\cite{journals/corr/BeattieLTWWKLGV16/Beattie/2016}, R2R
\cite{conf/cvpr/AndersonWTB0S0G18} and ALFRED \cite{conf/cvpr/ShridharTGBHMZF20}). Algorithmic
and deep imitation learning approaches for autonomous agents in these environments have been
proposed, but room for improvement
in both generalization performance and sample 
efficiency remains \cite{conf/aaai/ChenM11, conf/naacl/BiskYM16, conf/iclr/ShridharYCBTH21}.

% Generalization issues arise because 

The generalization
issue arises because there are many possible instructions or goals
that could be expressed with language and a learner
may not necessarily observe each one within its training
distribution. Some are ``out of distribution" and maintaining
performance on them is not guaranteed; a problem
well known the within reinforcement learning community \cite{journals/corr/abs-2111-09794/Kirk/2021}.
However, a peculiar feature of language instructions is that
language is \textit{compositional} in nature. This has
led to an interest in whether this aspect can be leveraged
to get better generalization on unseen goals made up
of familiar terms \cite{conf/icml/OhSLK17, journals/corr/HermannHGWFSSCJ17/Hermann/2017}.
However, even in simple environments such as BabyAI \cite{conf/iclr/Chevalier-Boisvert19},
and gSCAN \cite{conf/nips/RuisABBL20} this can still be difficult problem.

Various approaches to leveraging compositionality have been 
proposed, including gated word-channel attention 
\cite{conf/aaai/ChaplotSPRS18}, hierarchical processing
guided by parse-trees \cite{conf/emnlp/KuoKB21}, graph neural networks \cite{conf/ijcnlp/GaoHM20}, neural module networks \cite{conf/cvpr/2016/AndreasRDK15}, and extending agents with a boolean task algebra solver \cite{tasse2022generalisation}.
Closest to our approach are
\citet{journals/corr/abs-2009-13962/Heinze-Deml/2020, conf/icml/HanjieZN21}
which use attention to identify goal states, \citet{journals/jair/NarasimhanBJ18, journals/corr/abs-2202-10745/Ruis/2022}, which decompose goal identification and planning modules, \citet{conf/iclr/BahdanauHLHHKG19} which uses a discriminator to model reward for instructions and \citet{conf/TACL/Buch2021} which factorizes object classification over components.
We contribute
a new approach of learning sparse attention over factored observations, then attaching that attention module to a learned planning module. This can be shown to solve the compositional generalization problem by learning the correct correspondences between words and factors without spurious correlation.

\paragraph{Representation Sparsity} We hypothesize that sparsity 
is an important factor in the design of a compositional system
because it can bias the optimization procedure towards solutions
where relationships exist only between things that are actually
related and not just weakly correlated. Previous work has shown
that induced sparsity can improve both generalization \cite{conf/neurips/zhao2021a} and
model interpretability \cite{conf/icml/WongSM21}. Induced sparsity
has been applied both within the model weights \cite{conf/nips/JayakumarPROE20}
and also within the attention computation \cite{journals/taslp/ZhangZLZ19}.
In our work, we apply it in the space of all possible interactions
between words in the language and attributes of objects in the environment.

\paragraph{Sample Efficiency} In grounded language learning,
improved sample efficiency may enable
new use-cases, for example, the training of intelligent assistants by users who would not have the patience to give many
demonstrations of a desired behavior \cite{conf/aies/TuckerAD20}. Various tricks have been proposed to improve
sample efficiency in reinforcement learning in general \cite{conf/ijcai/Yu18},
including prioritized replay \cite{conf/aaai/HesselMHSODHPAS18}, data augmentation \cite{conf/nips/LaskinLSPAS20}
and model based learning or replay buffers \cite{conf/nips/HasseltHA19, conf/iclr/KaiserBMOCCEFKL20}.
Limited work exists on explicitly addressing sample
efficiency in the grounded language learning context \cite{conf/iclr/Chevalier-Boisvert19, journals/corr/abs-2007-12770/Hui/2020, conf/emnlp/QiuH0SS21}. In this work,
sample efficiency is one of our primary objectives and we
claim to achieve it using a functionally decomposed architecture
and offline learning.

% Talk about the learning environment here 
\section{Experimental Setup}
\label{sec:environment}
We study the performance of our proposed approach on the \texttt{GoToLocal} task of
the BabyAI environment. A detailed description
of the environment is given in Appendix \ref{sec:appendix_babyai}. The environment can be seen as a Goal-Conditioned Markov Decision Process,
(formally defined in \citet{conf/ijcai/Kaelbling93}).
Each episode is generated by a seed $i$ and has an initial state $s^{(i)}_0$. To obtain
a reward during an episode, the agent must successfully complete the language-encoded
instruction (denoted $g$) that it is given. The language is simple and generated by the use of a templating system.
\texttt{GoToLocal} consists only of statements ``go to (a|the) (color) (object)".
Each state is a fully observable 8-by-8 grid world and each cell (denoted $c_{ij}$) may contain an object, the agent, or nothing.

The information in each cell is \textit{disentangled}; the object's color is
in a separate channel to the object's type. We work with disentangled observations because they have been shown to improve the performance and sample-efficiency of attention-based models \citep[see, e.g.,][]{conf/icml/LoyndFcSH20}. This disentanglement is preserved by
embedding each component separately as factored embeddings $q_a$. The environment also comes with an
expert agent which can produce an optimal trajectory for a given initial
state and goal $\tau^{(i)}|s_0, g$.

The key performance metric is \textit{success rate}. A \textit{success} happens
if the agent completes the instruction within 64 steps.
We study compositional generalization and sample efficiency.

By \textit{compositional generalization} we mean maintaining performance when navigating
to objects with attribute combinations not seen during training.
To study this, we separate goals into $\mathcal{G}_{\text{ID}}$
and $\mathcal{G}_{\text{OOD}}$ following the principle of leaving one attribute combination out (shown in Table \ref{tab:dataset_split} and similar to the ``visual" split in \citet{conf/nips/RuisABBL20}). Then we create corresponding training
and validation datasets, $\mathcal{D}_{\text{train}}$, $\mathcal{D}_{\text{v\_ID}}$ and $\mathcal{D}_{\text{v\_OOD}}$ each containing
the same number of trajectories (10,000) per goal. Trajectories for each goal are generated in the same way, so we expect that a different split
of $\mathcal{G}_{\text{ID}}$ and $\mathcal{G}_{\text{OOD}}$ following the same principle will cause similar behavior in both the baselines and our models.
Finer details about the dataset construction are given in Appendix \ref{sec:appendix_dataset}.

\begin{table}[ht]
\resizebox{\linewidth}{!}{
\begin{tabular}{l|llllll}
\toprule
{}     & blue & red & green & yellow & purple & grey \\
box    & \cellcolor{blue!25} {} & \cellcolor{blue!25} {} & \cellcolor{red!25} {} & \cellcolor{blue!25} {} & \cellcolor{blue!25} {} & \cellcolor{red!25} {} \\
ball   & \cellcolor{blue!25} {} & \cellcolor{red!25} {} & \cellcolor{blue!25} {} & \cellcolor{blue!25} {} & \cellcolor{red!25} {} & \cellcolor{blue!25} {} \\
key    & \cellcolor{red!25} {} & \cellcolor{blue!25} {} & \cellcolor{blue!25} {} & \cellcolor{red!25} {} & \cellcolor{blue!25} {} & \cellcolor{blue!25} {} \\
\bottomrule
\end{tabular}
}
\caption{Split between $\mathcal{G}_{\text{ID}}$ and $\mathcal{G}_{\text{OOD}}$. Blue cells are object attributes appearing in
the goals for $\mathcal{G}_{\text{ID}}$ and red cells correspond to
those in $\mathcal{G}_{\text{OOD}}$.}
\label{tab:dataset_split}
\end{table}

By \textit{sample efficiency} we mean achieving a high level of performance
given a smaller number of samples than conventional methods might require. We denote $N$ as the number of
trajectories per goal that an agent has access to
and study performance at different levels of $N$. We train various models using $\mathcal{D}_{\text{train}}$
and describe the training methodology and results in Section \ref{sec:experiment_e2e}.

\begin{figure}[t]
\centering
\includegraphics[width=\linewidth, trim={0.3cm 1.3cm 0.3cm 1.3cm}, clip, keepaspectratio=1]{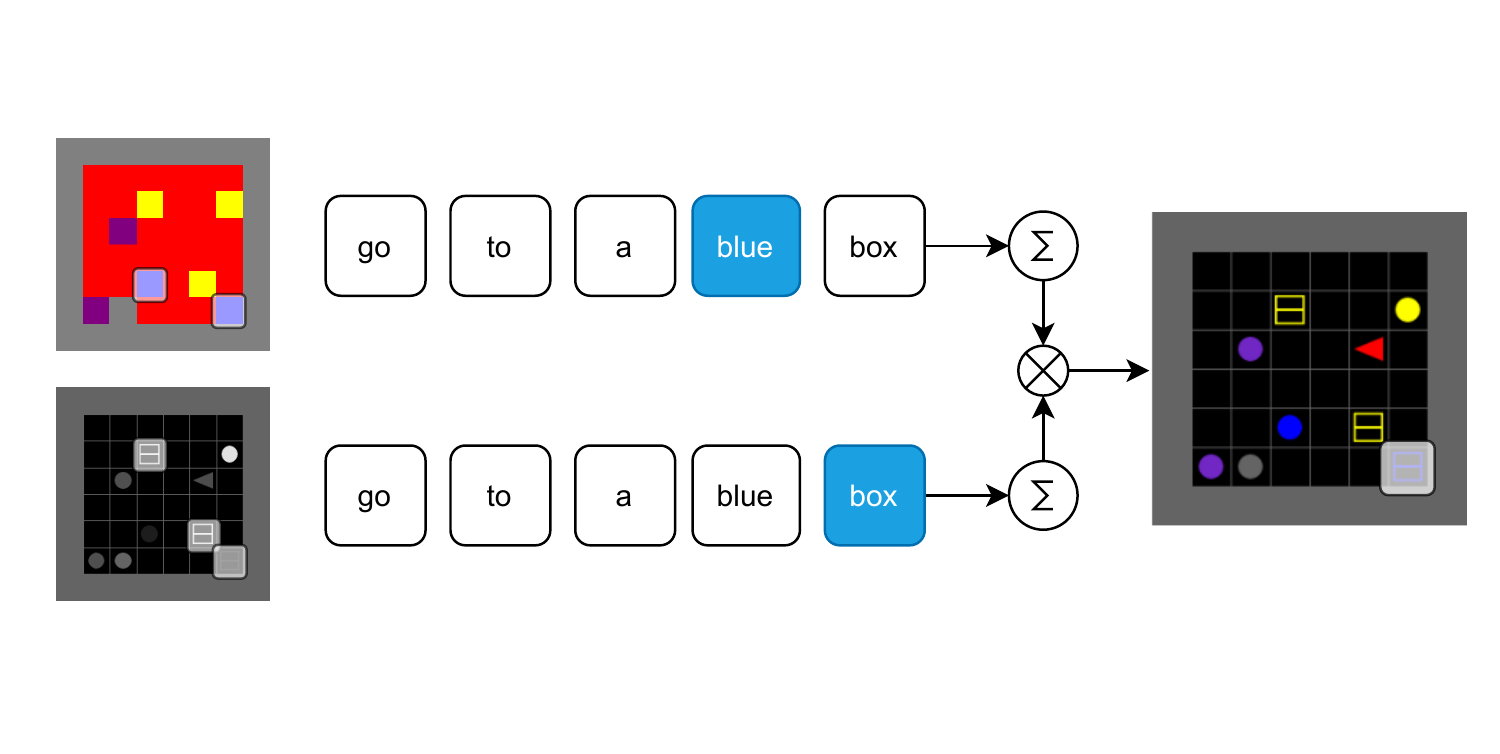}
\caption{Attention over separate components of the input representation. The model is a single layer of query-key attention applied to each component individually, where queries are image attribute values for a given component, the keys are the words and values are a one-tensor. Performing an AND relation on the components means taking the product of each attention operation.}
\label{fig:diagrams/cross_modal_attention.drawio.pdf}
\end{figure}

\section{Designing a learning method}
\label{sec:design}

We now design a learning agent with Section \ref{sec:related} in mind. 
To complete an instruction, the agent needs to identify the goal and plan actions to reach it. The
learning problem is decomposed into separate modules with separate training processes.
Subsections \ref{sec:sparse_architecture} and \ref{sec:discriminator} describe a sparse vision-language
architecture and training process for identifying goal cells ($S(s, g) \in \mathbb{R}^{H \times W})$.
Subsection \ref{sec:planning_vin} shows how to plan given that identification $\pi(a_t|S(s, g))$.

\begin{figure}[t]
\centering
\includegraphics[width=\linewidth, trim={0cm .5cm 0cm .8cm}, clip, keepaspectratio=1]{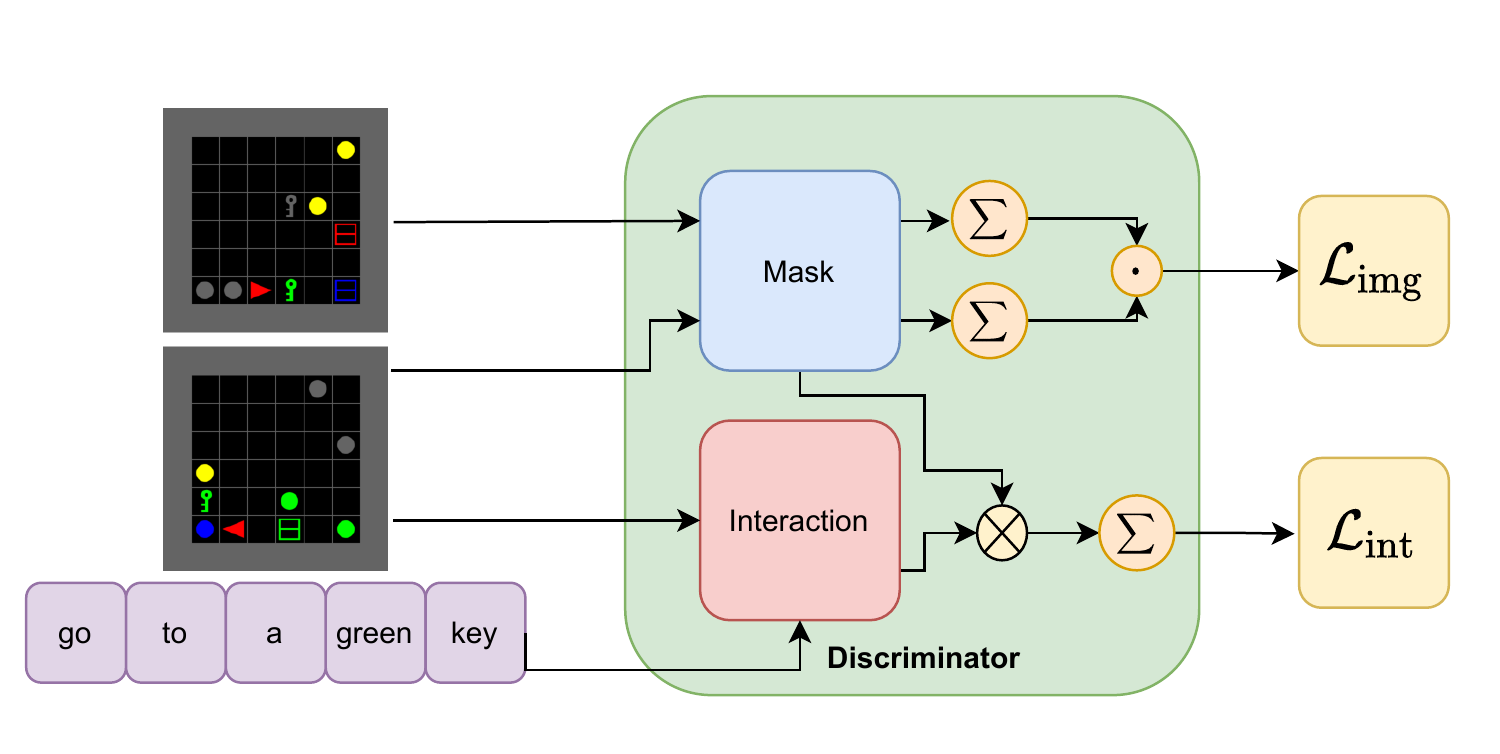}
\caption{Discriminator training method. $\mathcal{L}_{\text{img}}$ is used to train the ``mask" module. Because true examples are those where the agent is situated next to the same goal, an optimal mask module should select states the agent is facing. This can help with learning $S(s, g)$.}
\label{fig:diagrams/discriminator_architecture.drawio.pdf}
\end{figure}

\subsection{Sparse Factored Attention for Goal Identification}
\label{sec:sparse_architecture}

\begin{figure*}[ht]
    \centering
    \begin{subfigure}[b]{0.49\linewidth}
        \includegraphics[width=\textwidth]{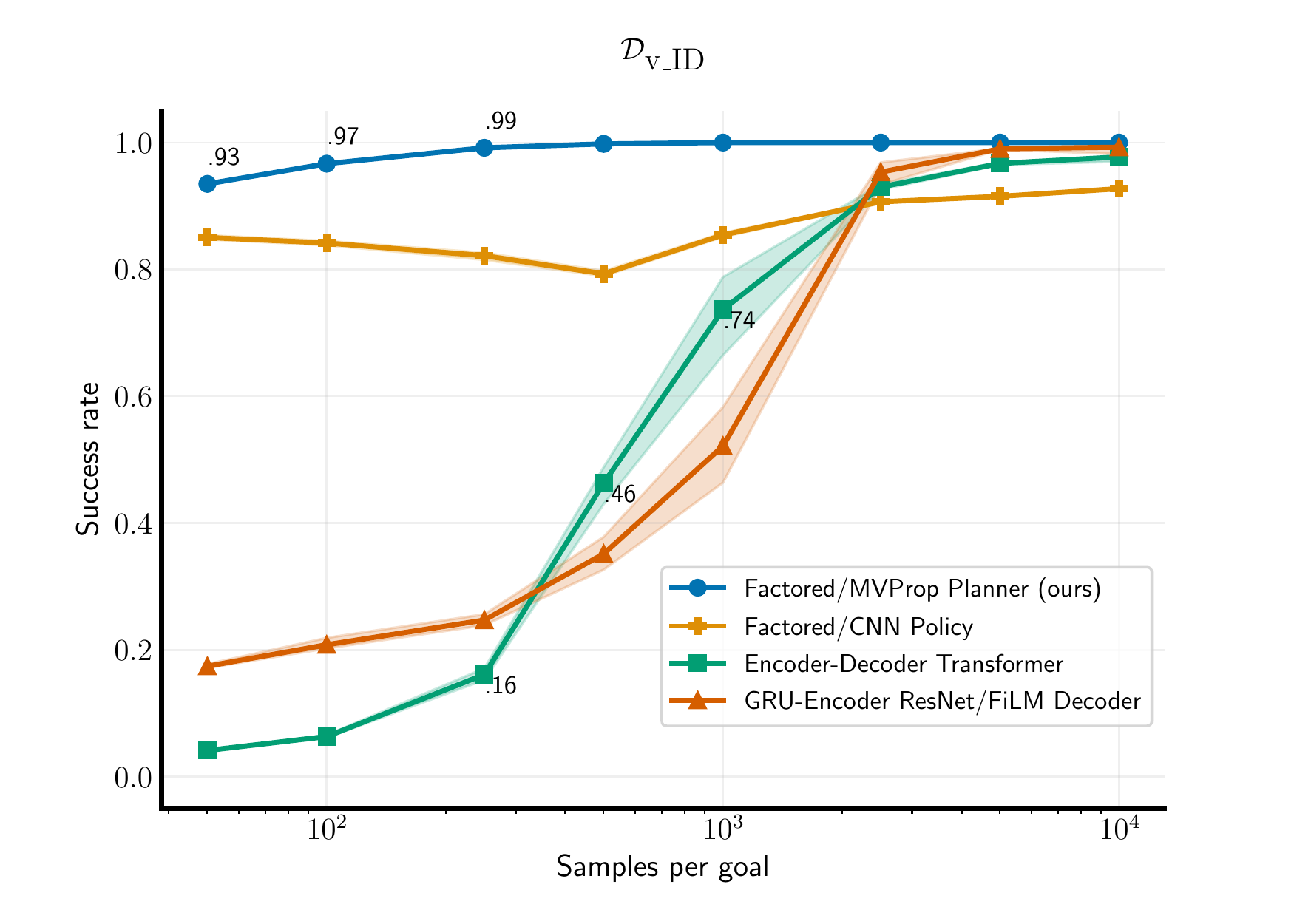}
    \end{subfigure}
    \hfill
    \begin{subfigure}[b]{0.49\linewidth}
        \includegraphics[width=\textwidth]{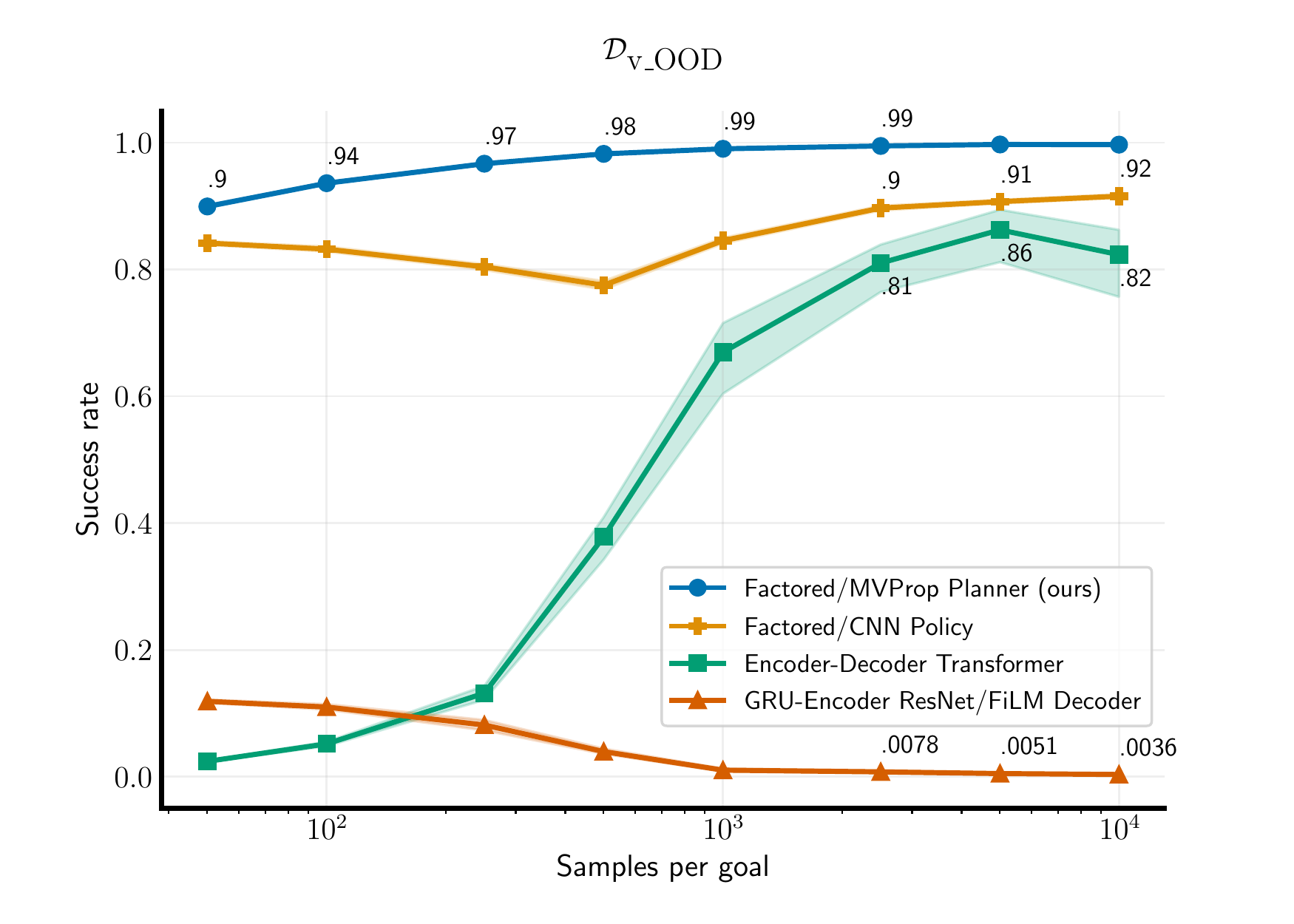}
    \end{subfigure}
    \caption{Success Rates on validation seeds. The x-axis is the log-scale number of samples per goal statement. Since there are 18 different goals in the training set, the total number of samples is $18 \times N$. Peak performance on within-distribution goals for prior methods in the same environment is typically reached at 2500 samples per goal, or 45,000 total samples. However, in the compositional generalization case ($\mathcal{D}_{\text{v\_OOD}}$), both baselines fail to maintain the same level of performance, although the Transformer baseline can provide a good amount of performance at a high number of samples. In comparison, Factored/MVProp (ours) reaches a comparable level of performance to peak performance of the baselines at 50 samples per goal, or 900 total samples, and maintains a consistent level of performance on the out-of-distribution validation set. Without a differentiable planner, Factored/CNN is still efficient but does not perform quite as well as Factored/MVProp.}
    \label{fig:plots/success_rates}
\end{figure*}
We hypothesize that learning to to match objects to descriptions by matching their factors to words individually is a process that generalizes more strongly
than matching all at once. For example, the agent should match ``red ball" to \texttt{red ball} because ``red" matches factor \texttt{red} and ``ball" matches \texttt{ball}. If the agent only learns that ``red ball" means \texttt{red ball}, as a whole, then it may not learn what the meaning of the parts are. Standard architectures, which can mix information between all the words or factors of the observation
might fall into the trap of doing the latter over the former. We propose two inductive biases to learn the former. 
The first bias is \textit{factor compositionality}. As language is a descriptive tool, words should operate at the level of
object properties and not entire objects. The second bias is \textit{sparsity} in
word/attribute relationships. A particular word should only match as many attributes as necessary.

From this intuition, we propose a ``Sparse Factored Attention" architecture,
pictured in Fig.~\ref{fig:diagrams/cross_modal_attention.drawio.pdf}.
The words are the keys and attributes are the queries.
However, a critical difference is that the attribute
embeddings for each $c_{jk}$ remain partitioned
into separate components $q_a$ corresponding to each factor. The normalized dot product ($\hat{c}_{jkq_a} \cdot \hat{g}_w$) is computed separately between
the instruction and the flattened observation cells for each factor, then the elementwise product is taken over each $q_a$:
\begin{equation}
    \label{eq:independent_attn}
    S(s_{t}, g)_{jk} = \prod_{q_a} \sigma (\alpha (\sum_w \hat{c}_{jkq_a} \cdot \hat{g}_w) + \beta)
\end{equation}
where $\alpha$ and $\beta$ are a single weight and bias applied to all dot product scores and $\sigma$ is the sigmoid
activation function. In practice, exp-sum-log is used in place of $\prod_{q_a}$ for training stability. To encourage sparsity
within the outer product, we add an L1 regularization penalty to the outer product of the normalized embedding spaces ($\lambda||\hat{E_c} \cdot \hat{E_w}^T||_1$) to the loss.
This goes beyond just penalizing $S(s, g)$; it ensures that the system's entire knowledge base is sparse, which
in turn assumes that no relationship exists between unseen pairs and is also not sensitive to imbalances in the
dataset regarding how often different objects appear in the observations.

\subsection{Training with a Discriminator}
\label{sec:discriminator}

We found that performance of end-to-end learning by differentiating through the
planner to our model was highly initialization sensitive. Instead we propose to learn goal-identification and planning separately. However, $\mathcal{D}$ does not have
labels of which cells are goal cells, but only full observations of the environment
at each step. To learn to identify the goals, we propose a self-supervised
objective in the form of a state-goal discriminator architecture $\hat D(s, g)$
shown in Fig.~\ref{fig:diagrams/discriminator_architecture.drawio.pdf}, which is 
trained to match end-states to their corresponding goals.

The discriminator is defined as:
\begin{equation}
    \label{eq:discriminator}
    \hat D(s, g) = \sum_{HW} M(s) \cdot S(s, g)
\end{equation}
where $S(s, g)$ is the trainable goal identification module and $M(s) \to \mathbb{R}^{H \times W}, \sum_{\text{HW}} M(s) = 1$ is a ``Mask Module".
The ``Mask Module" is a convolutional neural network with no downsampling or pooling and
returns a single-channel ``spatial softmax" with the same spatial dimensions as $s$.
Ideally the mask module should learn to identify the cell that the agent is facing.
When $M(s)$ and $S(s, g)$ are correctly learned, then $\hat{D}(s, g)$ answers whether the agent is
at the goal state. The training process for the discriminator uses a loss function similar to a triplet loss between positive,
negative, and anchor samples. Positive and negative goals are sampled from
the set of goals, then corresponding positive, anchor, and negative end-states.
Finer details of this process are given in Appendix \ref{sec:appendix_discriminator}.

\begin{figure*}[ht]
    \nocaption
    \centering
    \begin{subfigure}[t]{0.49\linewidth}
        \vspace{0pt}
        \includegraphics[width=\textwidth, keepaspectratio=1, trim={5cm 2.5cm 5cm 3.25cm}, clip]{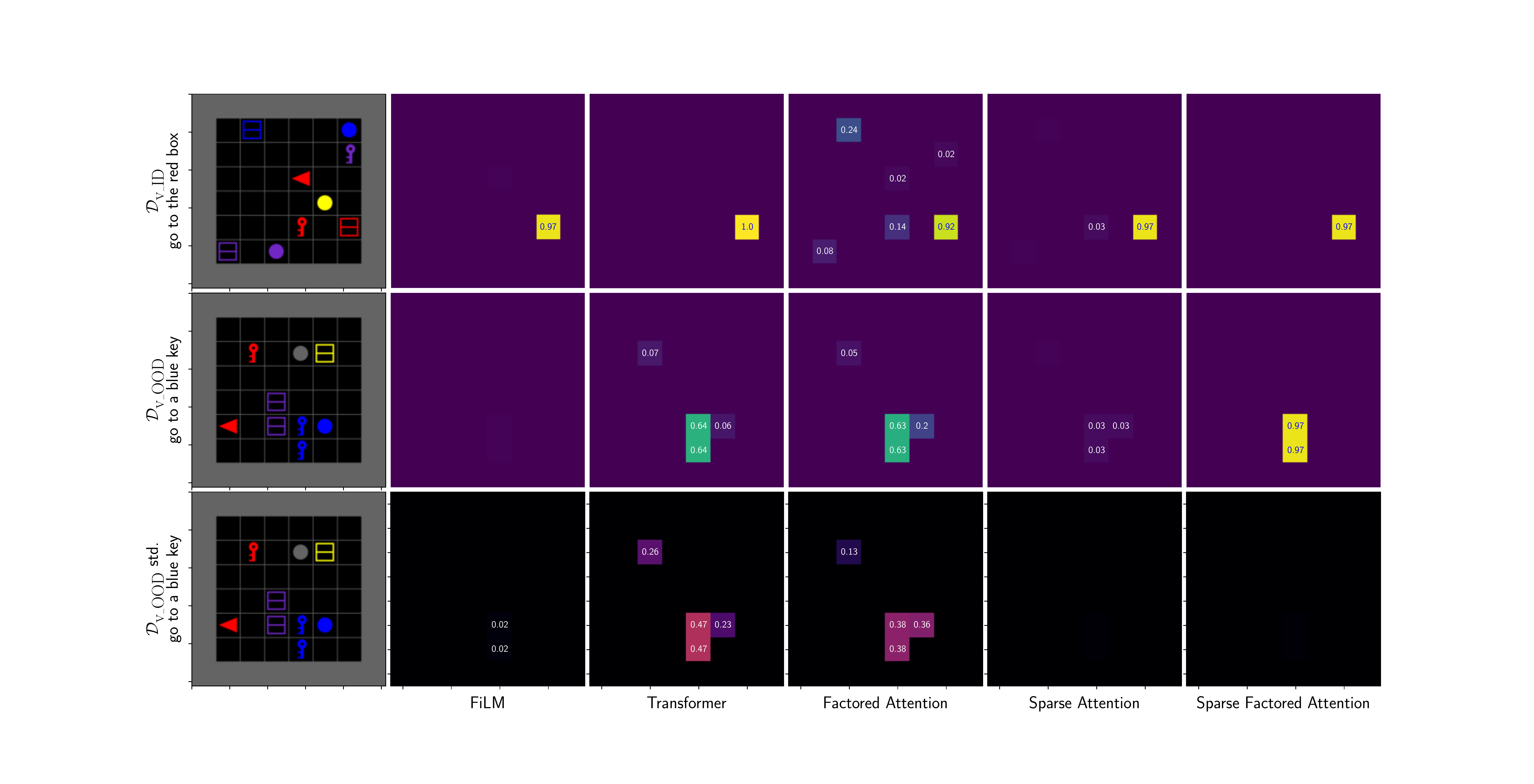}
        \caption{Qualitative evaluation of interaction networks on environment samples.
        The \textbf{top} row contains the mean activations a $\mathcal{D}_{\text{v\_ID}}$ sample, the \textbf{middle} and \textbf{bottom} rows are means and standard deviations on a $\mathcal{D}_{\text{v\_OOD}}$ sample. Other models either suffer from overfitting or high variance when predicting OOD goals.}
        \label{fig:plots/interaction_qualitative_evaluations.pdf}
        \end{subfigure}
    \hfill
    \begin{subfigure}[t]{0.49\linewidth}
        \vspace{0pt}
        \includegraphics[width=\textwidth, keepaspectratio=1, clip]{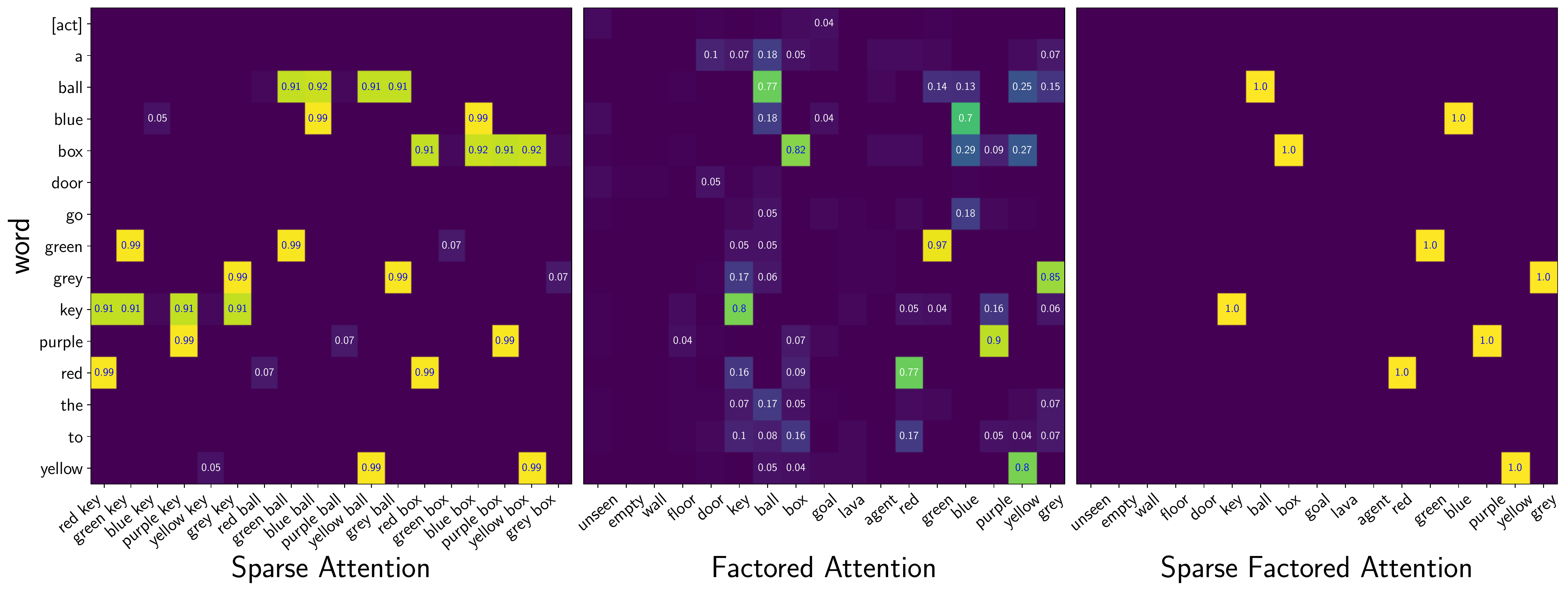}
        \caption{IQM of Embedding Internal Correlations for our method, showing the effect of applying L1 regularization to the embedding outer product. The horizontal axes correspond to factors and the vertical axes correspond to words. \textbf{Left:} when concatenating factor embeddings and applying sparse attention, unseen combinations such as \texttt{key/blue key} and \texttt{blue/blue key} are given little weight. \textbf{Middle:} without sparsity regularization, unrelated factors such as \texttt{box/yellow} are confused and less weight is given to the true correspondences. \textbf{Right:} ours, where the correspondences between words and factors are learned exactly and others are zero.}
        \label{fig:plots/independent_correlation_heatmaps.pdf}
    \end{subfigure}
\end{figure*}

\subsection{Planning Module}
\label{sec:planning_vin}

\begin{figure}[H]
\centering
\includegraphics[width=\linewidth, trim={0cm 0.4cm 0cm 0.8cm}, clip]{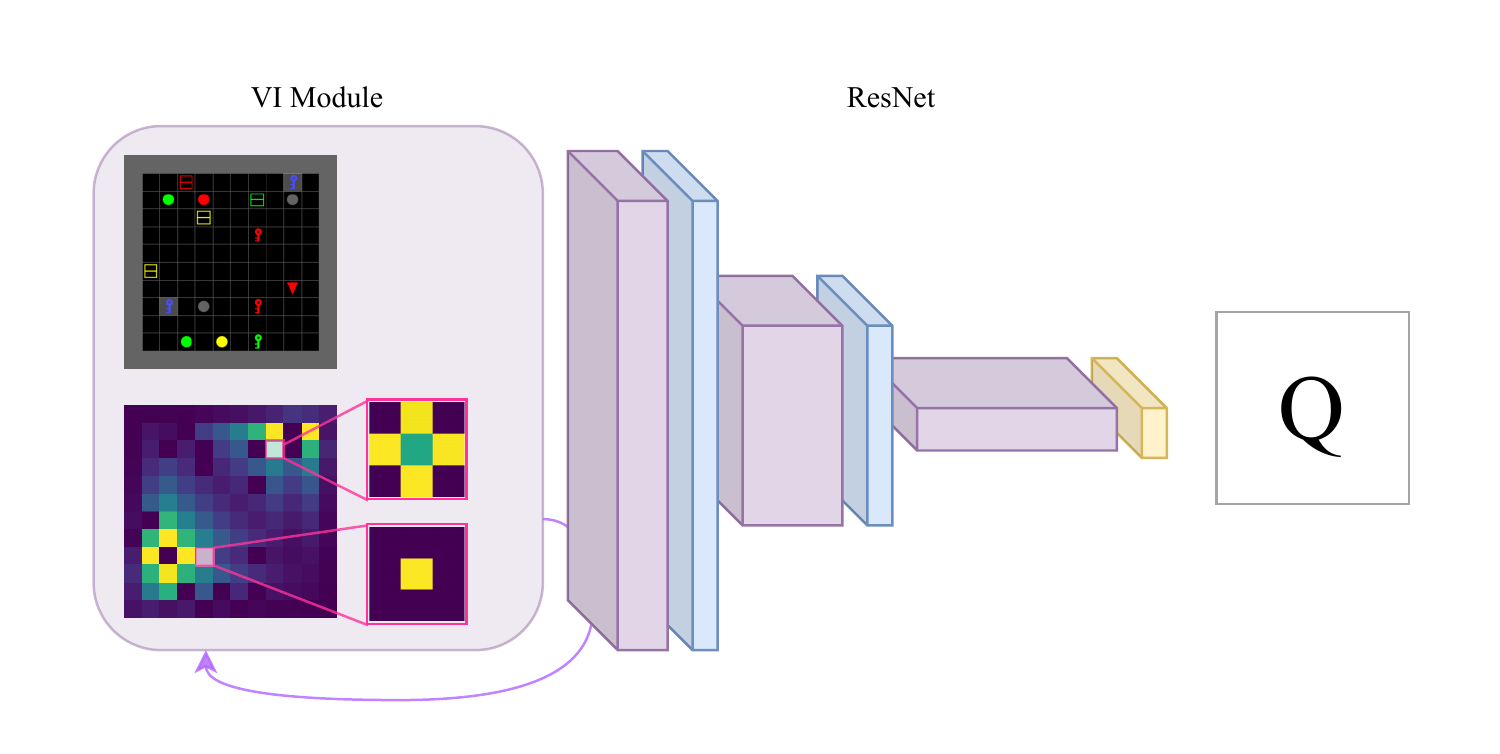}
\caption{Using a Value Propagation Network \cite{conf/iclr/NardelliSLKTU19} (VPN) to estimate the Q function. VPN is an extension of the Value Iteration Network \cite{conf/ijcai/TamarWTLA17} which makes the convolutional filter propagating value from one cell to its neighbors conditional on its inputs. The Q function is estimated by concatenating the output of the VPN with the estimated rewards, visual features, and agent state, then processing it with a ResNet.}
\label{fig:diagrams/vin_q_architecture.drawio.pdf}
\end{figure}

Once $S(s, g)$ is learned, with a knowledge of the connectivity between
cells, full observability of the environment, and the assumption that each
action moves the agent to a either the same cell or an adjacent, learning to plan
to reach a goal state becomes trivial. We extend Value Propagation Networks
\cite{conf/iclr/NardelliSLKTU19} for this purpose. Details of our implementation are given in Appendix \ref{sec:appendix_vpn}.

\section{Experimental results}

\subsection{End-to-End performance on the benchmark task}
\label{sec:experiment_e2e}
We first examine performance and sample efficiency on both $\mathcal{D}_{\text{v\_ID}}$ and $\mathcal{D}_{\text{v\_OOD}}$
using the experimental setup described in Section \ref{sec:environment}. We train our approach and several baseline models for the same number (70,000) of training
steps over many values of $N$ and 10 random intializations. The models are briefly described as follows:
\paragraph{Factored/MVProp (blue circles, ours)} Sparse Factored Attention is pre-trained with the Discriminator in Section \ref{sec:discriminator} and frozen, then we only learn the planning and value networks in Section \ref{sec:planning_vin}.
\paragraph{Factored/CNN (light orange plus marks)} Ablation of our model with a skipped planning step; detected goals and observations are processed directly into a policy using a convolutional network. 
\paragraph{Transformer (green squares)} Standard encoder-decoder transformer, encoder inputs are position-encoded instruction word embeddings, decoder inputs are position-encoded flattened cells and a [CLS] token used to predict the policy.
\paragraph{GRU-Encoder ResNet/FiLM Decoder (red triangles)} Process visual observation into policy with interleaved FiLM conditioning on the GRU-encoded instruction, similar to \citet{journals/corr/abs-2007-12770/Hui/2020}.

The training objective is behavioral cloning of the expert policy.  The model is evaluated is every
500 steps. Evaluation is performed in a running copy of the environment seeded using each of the stored seeds in
the validation sets. To succeed the agent must solve the task - it is not enough to copy what
the expert does on most steps. Further details are given in Appendices \ref{sec:appendix_baselines} and \ref{sec:appendix_method_e2e}.

In contrast to both baselines, our method in Fig.~\ref{fig:plots/success_rates} attains a high level of performance on both
$\mathcal{D}_{\text{v\_{ID}}}$ and $\mathcal{D}_{\text{v\_{OOD}}}$, even with a small number of samples, significantly outperforming both baselines even when those models have
a greater number of samples available to learn from.

\subsection{Examination of Interaction Module Architectures}
\label{sec:qualitative_examination}

We also examine what it is about our model architecture that explains its performance on the benchmark task. We perform an ablation study to examine the effectiveness of different architectures for $S(s, g)$. Performance is measured
using a ``soft $F_1$ score" against a ground truth on goal locations, as this is essentially
an imbalanced classification problem. The metric is described in more detail in Appendix \ref{sec:appendix_soft_f1_score}

\begin{table}[ht]
\resizebox{\linewidth}{!}{
\begin{tabular}{lll}
\toprule
{} & $\mathcal{D}_{\text{v\_ID}}$ & $\mathcal{D}_{\text{v\_OOD}}$ \\
\midrule
FiLM \cite{conf/aaai/PerezSVDC18}  &   0.983 ± 0.000 &       0.015 ± 0.004 \\
Transformer \cite{conf/nips/VaswaniSPUJGKP17}  &   1.000 ± 0.000 &       0.799 ± 0.028 \\
\hline
Sparse Attention                   &   0.974 ± 0.000 &       0.069 ± 0.001 \\
Factored Attention              &   0.891 ± 0.015 &       0.739 ± 0.028 \\
\textbf{Sparse Factored Attention}       &   0.951 ± 0.000 &       0.951 ± 0.000 \\
\bottomrule
\end{tabular}
}
\caption{Inter-quartile mean (IQM) of soft F1 scores (predicted goal location versus ground truth goal location) across seeds, dataset sizes, and checkpoints, with added
95\% confidence intervals. Sparse Factored Attention scores consistently well on both datasets.}
\label{tab:scores_interaction}
\end{table}

Each architecture for $S(s, g)$ was trained using $\mathcal{D}_{\text{train}}$ for 200,000 iterations with the parameters in Appendix \ref{sec:appendix_training_s}.
The IQM and 95\% confidence interval across seeds and top-10 checkpoints
are reported in Table \ref{tab:scores_interaction} using the package and method provided by \cite{journals/corr/abs-2108-13264/Agarwal/2021}.
While not perfect, our Sparse Factored Attention model achieves high $F_1$ scores
both $\mathcal{D}_{\text{v\_ID}}$ and $\mathcal{D}_{\text{v\_OOD}}$.

We also visualize mean model predictions and their variance across initializations on sample datapoints from
both $\mathcal{D}_{\text{v\_ID}}$ and $\mathcal{D}_{\text{v\_OOD}}$ in Fig.~\ref{fig:plots/interaction_qualitative_evaluations.pdf}.
The average is over instances with $F_1$ scores in the upper 75\% range for their class. FiLM and Sparse Attention fail to identify
the test-set goal, and the Transformer and Factored Attention models exhibit high variance on $\mathcal{D}_{\text{v\_OOD}}$ between initializations.
Only our Sparse Factored Attention model reliably identifies the goal on both datasets.

\subsection{Qualitative Evaluation of Model Weights}

Since the Factored Attention model is very simple and its only parameters
are the embeddings and single weight and bias, we can also visualize
``what the model has learned" by taking the mean normalized outer product of
both attribute $E_c$ and word $E_w$ embeddings for models
shown in Fig.~\ref{fig:plots/independent_correlation_heatmaps.pdf}. A perfect learner should 
learn a sparse correspondence between each attribute and its corresponding word;
it should not confound attributes of different types. The heatmaps show the importance
of sparsity regularization on the outer product of the embeddings. Without
sparsity regularization, the mean correlation between a word and its correct attribute is
weaker and not consistent across all initializations.
There are also other ``unwanted" confounding correlations, for example, between ``box"
and \texttt{blue}, which also appear more strongly in some initialization and data limit combinations
as indicated by its high standard deviation. In contrast, the Sparse Factored Attention model
displays an almost perfect correlation between each word and the corresponding attribute
and very little variance between checkpoints (not pictured). In this sense, we can
be much more confident that the Sparse Factored Attention model has \textit{actually
learned the symbol grounding} and the meaning of the words as they relate to cell attributes
in the environment.

\section{Conclusion}
\label{sec:conclusion}

We studied the problem of compositional generalization and sample efficient
grounded language learning for a vision-language navigation agent. We showed that
even under strong assumptions on environment conditions such as full observability
and disentanglement of inputs, compositional generalization and
sample efficiency do not arise automatically with standard learning
approaches. We demonstrate how such conditions can be leveraged by our Sparse Factored Attention model presented in Section \ref{sec:sparse_architecture}.
We demonstrate a method to learn goal identification without labels in Section \ref{sec:discriminator}
and planning Section \ref{sec:planning_vin} using a small number of offline trajectories.
We further showed superior sample efficiency and generalization performance
in Section \ref{sec:experiment_e2e} and perform a model analysis and ablation study in
Section \ref{sec:qualitative_examination} to show how our proposed approach works the
way we intended.

\section{Limitations of this Work}
\label{sec:limitations}

\paragraph{Goal identification and planning}
The goal identification and planning methods proposed
in Section \ref{sec:planning_vin} do not work over compound goals.
The discriminator training method in Section \ref{sec:discriminator}
requires that $\mathcal{D}_{\text{train}}$ can be partitioned
into subsets corresponding to each goal and that there is
at most a many-to-one relationship between goal cell configurations
and language statements.

\paragraph{Measuring sample efficiency}
Testing sample efficiency of gradient-based methods learned from off-policy datasets
is not a well specified problem, since each training step could be
used to improve the model performance by a small amount an arbitrary number
of times. It was a qualitative judgment of the researchers of when to stop
training, and we used the same upper bound on training steps for all models
to ensure a fair comparison.

Further limitations of this work are discussed in Appendix \ref{sec:appendix_limitations}.

\section{Responsible Research Statement}
\label{sec:statement}

We also provide details regarding code and reproducibility in Appendix \ref{sec:appendix_reproducibility}
and computational resource usage in Appendix \ref{sec:appendix_resource_usage}. We do not anticipate any special ethical issues to arise from this work as it is foundational in nature and uses a synthetically generated dataset. However, the methods presented in this work may be more
amenable to analytic languages as opposed to synthetic ones.

\section{Acknowledgements}
We thank Yonatan Bisk for his valuable feedback and suggestions on this work. We also acknowledge
the computational resources provided by the Aalto Science-IT project and the support within the Academy of Finland Flagship programme: Finnish Center for Artificial Intelligence (FCAI).

\bibliography{anthology,bibliography}

%\newpage
\vfill\null
\newpage

\appendix

\section{Details of the BabyAI Environment}
\label{sec:appendix_babyai}
\begin{figure}[ht]
    \centering
    \includegraphics[width=\linewidth]{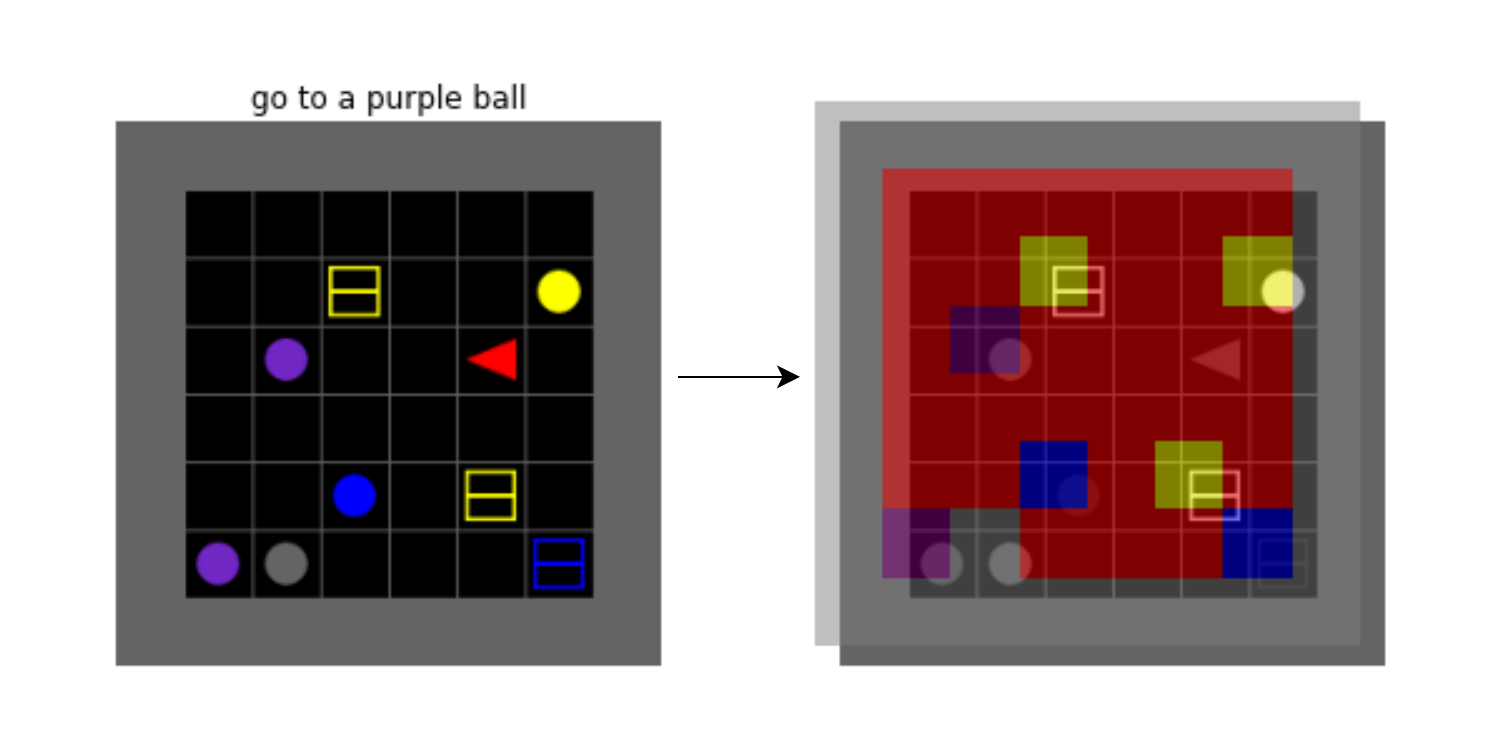}
    \caption{An illustration of the integer-encoded inputs provided by the BabyAI environment. Color and shape information are encoded in separate channels and are independent from each other.}
    \label{fig:diagrams/image_with_component_channels.drawio.pdf}
\end{figure}

BabyAI is a simple grid world-like environment based on Minigrid \cite{gym_minigrid}.
chose to use this environment for this project due to its simplicity, ease of
generating expert trajectories, and input representation characteristics.
In the environment, the agent is given instructions to complete in a sythetically generated
language that is a subset of English. The seed for the environment, $i$,
determines its initial state $s_0$ and goal $g$, which comes from the set $\mathcal{G}$ for a given level.
Within the environment, there are
a few  different object types (\texttt{ball}, \texttt{box}, \texttt{key}) each of which
may be one of six different colors (\texttt{red}, \texttt{blue}, \texttt{green}, \texttt{grey}, \texttt{purple}, \texttt{yellow}). The
agent can face one of four different directions. There are seven actions available to the 
agent: \texttt{turn left}, \texttt{turn right}, \texttt{go forward}, \texttt{open}, \texttt{pick up}, \texttt{put down} and \texttt{signal done}. The 
original implementation provides partial observations, however we modify the
environment to make the state space fully observable due to the inherent difficulty
planning over unobservable states.\footnote{We also reproduce
the relevant experiments in \cite{conf/iclr/Chevalier-Boisvert19} using this fully-observable state space for
fair comparison in Section \ref{sec:experiment_e2e} of this work.} The observations
are subdivided into cells as explained in Section \ref{sec:environment}. Each
cell is a disentangled vector of integers of comprised of three components,
the first corresponding to the object type,
the second corresponding to the color and the third corresponding to the object that
the agent is holding.

The goals $g$ come in the form of simple language statements
such as ``go to a red box". BabyAI comes in several ``levels". Each level requires
the agent to demonstrate competency at a certain subset of ``skills",
summarized in Table 1 of the original by \citet{conf/iclr/Chevalier-Boisvert19}.

In this work, we focus on the \texttt{GoToLocal} task, where the agent
must learn to reach the goal object indicated in the language-encoded instruction
by navigating to the correct location in an $8 \times 8$ grid world and then performing the \texttt{signal done} action within a
fixed number of steps. Performing \texttt{signal done} facing the wrong cell terminates
the episode with a reward of zero. Requiring the \texttt{signal done} action precludes the trivial
solution of ignoring $g$ and visiting every object until successful. Other objects may exist in the grid
as \textit{distractors}; non-goal objects that the agent must learn to ignore and navigate
around depending on the goal.

\section{Collecting Trajectories for the Dataset}
\label{sec:appendix_dataset}

In the \texttt{GoToLocal} task there are 36 possible goal statements.
Each statement begins with ``go to", followed by ``the" or ``a", then color and object terms. To collect the seeds
to generate each environment and their corresponding solutions $\tau_i|s_0, g$,
we iterate consecutively through random seeds starting
from zero and reset the environment using each seed. The environment is ``solved"
using the provided \texttt{BotAgent}, which implements an optimal policy.
We do not want our measurements or training
to be biased by imbalances in the dataset, so we want to ensure that each
goal has the same number of samples in $\mathcal{D}$. 10,000 state-action trajectories with a length of
at least 7 are stored for each goal $g$. A trajectory $\tau$ is a tuple
$(x, (s_0, ..., s_t), (a_0, ..., a_t), (r_0, ..., r_t), g)$,
consisting of (respectively), the seed, state trajectory, action trajectory, rewards and goal.

We split the data into training, ``in-distribution" and ``combinatorial generalization"
(out of distribution) validation sets. To make these splits, we first split the
goals into ``in-distribution" goals $\mathcal{G}_{\text{ID}}$ and ``combinatorial generalization" goals $\mathcal{G}_{\text{OOD}}$. One color and object combination is omitted
from $\mathcal{G}_{\text{ID}}$ for each color and placed in $\mathcal{G}_{\text{OOD}}$,
specifically, goals containing
\texttt{red ball}, \texttt{green box}, \texttt{blue key},
\texttt{purple ball}, \texttt{grey box} and 
\texttt{yellow key}.
The ``in-distribution" validation set $\mathcal{D}_{\text{v\_ID}}$ consists of
the last 20 trajectories in $\mathcal{D}$ corresponding to each $g \in \mathcal{G}_{\text{ID}}$. The ``combinatorial generalization" set $\mathcal{D}_{\text{v\_OOD}}$ is defined similarly with the last 40 trajectories in $\mathcal{G}_{\text{OOD}}$.\footnote{The reason for using the last 40 trajectories
is to ensure that both validation datasets have the same number trajectories in total;
since there are twice as many goals covered in $\mathcal{D}_{\text{v\_ID}}$} The training set $\mathcal{D}$
consists of all trajectories corresponding to $g \in \mathcal{G}_{\text{ID}}$,
excluding those in $\mathcal{D}_{\text{v\_ID}}$.

\section{Details of the Baselines}
\label{sec:appendix_baselines}

The first baseline is similar to the architecture
used in \cite{journals/corr/abs-2007-12770/Hui/2020}; featuring a GRU to encode $g$,
a ResNet to encode $s$ and the use of FiLM layers \cite{conf/aaai/PerezSVDC18} to
modulate feature maps according to the encoded $g$, which in turn is flattened and concatenated
with an embedding corresponding to the agent's current direction to produce
a hidden representation $z$. The policy $\pi$ is estimated using an MLP from $z$.
The only difference to \cite{journals/corr/abs-2007-12770/Hui/2020}
is that the memory module used to handle partial observability and exploration is removed,
since the environment is fully observable.

The second baseline is an encoder-decoder Transformer
model \cite{conf/nips/VaswaniSPUJGKP17}, where the input sequence is the
individual words in $g$ added with their 1D positional encodings, and the output sequence
is the 2D encoded observation $s$ added with their 2D positional encodings. A
classification token is appended to the end of the output sequence, which uses a linear prediction head to estimate $\pi$ in the same way as above.
10000 steps of learning rate warmup followed by subsequent logarithmic decay in the learning rate are used when training the Transformer.

For all models, an embedding dimension of 32 is used
for both the words in $g$ and each attribute in $c_{jk}$, implying that the total embedding
dimension is 96 after each embedded attribute is concatenated together. The batch size
and learning rate for Adam used during training are 32 and $10^{-4}$ respectively.

\section{Training the Discriminator}
\label{sec:appendix_discriminator}
Two goals, $g_{+}, g_{-}$ are sampled without replacement uniformly from the set of all known goals $\mathcal{G_{\text{v\_ID}}}$.
Two trajectories are sampled without replacement from
$\{\mathcal{D}_{\text{train}}| g = g_{+}\}$, $\tau^{g_{+}}_1, \tau^{g_{+}}_2$ 
and one trajectory is sampled from $\{\mathcal{D}_{\text{train}} | g = g_{-}\}$, $\tau^{g_{-}}$.
$s_r$ is assumed to be the rewarding
states for all three trajectories and are denoted $(s_r^{g_{+}})_1, (s_r^{g_{+}})_2, (s_r^{g_{-}})_1$.
With probability $\frac{1}{|\mathcal{G}|}$, $(s_r^{g_{-}})_1$ is replaced with a random state
in $\tau_{0:T - 1}^{g_{-}}$, so that the discriminator also sees states that are
not rewarding for any goal. The discriminator's inputs and labels are tuples $(s_1, s_2, g, y)$.
In this tuple, $s_1$ is an "anchor" state, $s_2$ is a comparison state, $g$ is the goal and $y$ is the label.
The tuple $((s_r^{g_{+}})_1, (s_r^{g_{+}})_2, g_{+}, 1)$ is a ``true" example and
the tuple $((s_r^{g_{+}})_1, (s_r^{g_{-}})_1, g_{+}, 0)$ is a ``false" example. True and false
examples are sampled consecutively.

We define the loss for the discriminator as:
\begin{equation}
\label{eq:discriminator_loss}
\resizebox{0.85\linewidth}{!}{
$\mathcal{L}_{D}(s_1, s_2, g, y) = \mathcal{L}_{\textrm{int}}(s_2, g, y) + \mathcal{L}_{\textrm{img}}(s_1, s_2, y)$
}
\end{equation}
The ``interaction loss" $\mathcal{L}_{\text{int}}$ is used to optimize
$S(s, g)$. As $S$ classifies whether a given $s$ is a rewarding state for $g$, the loss
is a binary-cross-entropy loss, where the outputs of $S$ are logits:
\begin{equation}
\resizebox{0.85\linewidth}{!}{
$\mathcal{L}_{\textrm{int}}(s_2, g, y) = y \log D(s_2, g) + (1 - y) \log (1 - D(s_2, g))$
}
\end{equation}
The image-matching loss $\mathcal{L}_{\text{img}}$ is used to resolve the ambiguity of whether
a high loss value in $\mathcal{L}_{\text{int}}$ was caused by an incorrect parameterization of
$M(s)$ or $S(s, g)$. Define the \textit{mask-weighted image} as $I(s) = \sum_{\text{HW}} M(s) \odot s$
and the \textit{normalized mask-weighted image} as $\hat{I}(s) = \frac{I(s)}{||I(s)||^2_2}$
Then the normalized image-matching loss $L_{\textrm{img}}$ is given by:\footnote{We use mean-squared error as opposed to binary cross entropy loss for the the image-matching loss as we found that in practice it was less
sensitive to label noise, which was present in this problem, since goals such as ``go to a red key"
and ``go to the red key" involve the same object color combination but are nevertheless treated as separate goals
by the discriminator.}
\begin{equation}
\label{eq:img_loss}
\resizebox{0.85\linewidth}{!}{
$\mathcal{L}_{\textrm{img}} (s_1, s_2, y) = ||(\hat {I}(s_1) \cdot \hat {I}(s_2)) -  y||^2_2$
}
\end{equation}

\section{Planning with Value Iteration}
\label{sec:appendix_vpn}

Value-based differentiable planning networks assume the existence of a function $r(s, g) : \mathbb{R}^{H \times W \times A}$
which returns the cell-action combinations in $s$ that give a reward for being reached by an agent. In this case, $r$ is modelling
a reward function for goal $g$ in terms of $c_{jk}$. Knowing both this function and the dynamics $p(s_{t + 1}|s, a_t)$
with a discrete state space enables using \textit{Value Iteration} \cite{bellman1957markovian} to solve
for the \textit{optimal value function} $V^*$, which induces an \textit{optimal policy}:
\begin{equation}
\label{eq:optimal_pi}
\resizebox{0.85\linewidth}{!}{
    $\pi^* = \max_{a} Q(s, a) = \max_{a} \sum_{a \in |\mathcal{A}} r(s, a) + \gamma p(s_{t + 1}|s, a_t) V(s_{t + 1})$
}
\end{equation}
In this case, we do not know the dynamics exactly, but we have a
prior that we can start from, which is that all neighboring cells to a given cell
are uniformly reachable from the current cell by any action $p(c_{j + l, k + m}|a_t, c_{jk}), l, m \in [-1, 1], a \in \mathcal{A}$.
In this problem, the agent's occupancy of a cell $c_{jk}$ corresponds to a state $s$ given the initialization $s_0$,
so a mapping exists from values of cells to values of states up to the agent's rotation
given an initialization $V(c_{jk}) \to V(s|s_0)$.
\iffalse
We also know that the action-value function $Q(s, a_t)$ is related to the value function $V(s)$
by the dynamics:
\begin{equation}
    \label{eq:q_function_defintion}
    \resizebox{0.85\linewidth}{!}{
$\begin{aligned}
    Q(s, a_t, g) = \sum\limits_{s_{t + 1} \in \mathcal{S}} &p(s_{t + 1}|s_{t}, a_t) [r(s, a_t, g, s_{t + 1}) +\\
                                                             &\gamma V(s_{t + 1}, g)]
\end{aligned}$
    }
\end{equation}
where $\gamma$ is the discount factor used to estimate the return of $s$ under $g$: $R(s, g) = \sum \gamma^k r_{t + k}$.
\fi

To refine our estimate of the the dynamics $p(s_{t + 1}|s, a_t)$ and improve our estimate of $Q(s, a_t, g)$,
we can use the above assumptions and a differentiable planning method known as a \textit{Value Iteration Network} (VIN) \cite{conf/ijcai/TamarWTLA17}. Starting with
$V_0(c_{jk}) = r(c_{jk}, g)$, VIN re-expresses
value-iteration as a form of \textit{convolution} performed recursively $K$ times:
\begin{equation}
\resizebox{0.85\linewidth}{!}{
  $V_{k + 1}(c_{jk}, g) = \max \begin{cases}
    V_k(c_{jk}, g), \\
    \max\limits_{a \in \mathcal{A}} \sum\limits_{l, m \in \mathcal{N}(c_{jk})} \mathbf{P}_{a, l - j, m - k} V_k(c_{lm}, g)
  \end{cases}$
}
\end{equation}
where $\mathcal{N}(c_{jk})$ are the neighbors of a cell and $\mathbf{P}$ is a learnable
linear estimate of the dynamics (the transition probabilities to neighboring cells for each action). In reality,
the dynamics are dependent on what the neighboring cells actually contain.
\textit{Max Value Propagation Networks} (MVProp) \cite{conf/iclr/NardelliSLKTU19} extend on VIN by replacing
$\mathbf{P}$ with a scalar \textit{propagation weight} conditioned on the current cell $\phi(c_{jk})$,
where $\phi$ is any learnable function with non-negative output. In that sense, we learn to model
how value \textit{propagates} around the cells. Using the dataset $\mathcal{D}$ we can generate traces of returns from
trajectories using an optimal planner with discount factor $\gamma$. Then learning $Q(s, a_t, g)$
is done by minimizing the empirical risk with respect to some loss function $\mathcal{L}$:
\begin{equation}
\label{eq:min_q_loss}
\resizebox{0.88\linewidth}{!}{
  $\arg \min_{Q_{\theta}} \mathbb{E}_{s, a_t \sim \mathcal{D_{\text{tr}}}} \mathcal{L} (Q(s, a_t, g), R(s, a_t)))$
 }
\end{equation}
In the MVProp framework, it is the responsibility of the consumer of $V_K(s, g)$ to map neighboring values
of a cell to Q values for actions. Both \citet{conf/ijcai/TamarWTLA17} and \citet{conf/iclr/NardelliSLKTU19} resolve
this problem by including the cell that the agent is currently occupying as part of
the state. However, this information is not available to us in $\mathcal{D}$ as we have only the state $s$ and action 
observation $a_t$. In practice, this problem turns out not to be insurmountable and good performance can be achieved
by simply concatenating as additional channels $V_0(s, g)$ and
$V_k(s, g)$ to the initial encoding of
$s$ and using a Convolutional Neural Network to encode the image into a single vector of
which represents the vector-valued output $Q(s, g) \to \mathbb{R}^{|\mathcal{A}|}$, eg the action-value
function for all actions.

Finally, there is the question of which loss function to use to learn $Q(s, a_t, g)$.
We observed that simply using mean-squared error loss between $R(s, a_t)$ and
$Q(s, a_t, g)$ led to over-optimistic estimates of Q-values for non-chosen actions.
To fix this problem, we added an additional term penalizing any non-zero
value for those actions: similar to Conservative Q Learning \cite{conf/nips/KumarZTL20}:
\begin{equation}
\resizebox{0.85\linewidth}{!}{
$\begin{aligned}
    \mathcal{L}_{\text{VIN}}(s, a_t, g) =& \ || R(s, a_t, g) - Q(s, a_t, g) ||^2_2 + \\ & \lambda || Q(s, a_{-}, g), {a_{-} \in \{\mathcal{A} \setminus a_t\}} ||^2_2
\end{aligned}$
}
\end{equation}

\section{Training Parameters of $S(s, g)$}
\label{sec:appendix_training_s}
$S(s, g)$ is trained for 200,000 steps, using a learning rate of $10^{-5}$, a batch size of 1024 and
16-bit mixed precision used for the model weights and embeddings. During training, models were evaluated both $\mathcal{D}_{\text{v\_ID}}$
and $\mathcal{D}_{\text{v\_OOD}}$ every 20 training steps. The top-10 performing model checkpoints
by $F_1$ score on $\mathcal{D}_{\text{v\_ID}}$ were stored, along with their $F_1$ score on $\mathcal{D}_{\text{v\_OOD}}$.

\section{Soft F1 Score}
\label{sec:appendix_soft_f1_score}
The problem in Section \ref{sec:discriminator} is unbalanced; there are a small number of goal states
and a large number of non-goal states. Therefore, we propose to use a metric that is robust to the class imbalance,
but also takes into account the weight of the predictions as this will be used as the
reward model in the planner. The metric is a ``soft F1 score" is defined
as the harmonic mean of soft-precision and soft-recall, for a single trajectory $i$ (with indexes omitted for brevity):
\begin{equation}
\label{eq:metric}
\resizebox{0.80\linewidth}{!}{$%
\begin{aligned}
P &= \frac{\sum_{\text{HW}}^{jk} y_{jk} S(s, g)_{jk}}{\sum_{\text{HW}}^{jk} (y_{jk} S(s, g)_{jk} + (1 - y_{jk})S(s, g)_{jk})} \\
R &= \sum_{\text{HW}}^{jk} y_{jk} S(s, g)_{jk} / \sum_{\text{HW}}^{jk} (y_{jk}) \\
F_1 &= 2PR/(P + R)
\end{aligned}
$}
\end{equation}
A high value of soft-$F_1$ indicates that both precision \textit{and} recall are high.

\section{End-to-end usage our proposed model}
\label{sec:appendix_method_e2e}
The model is trained in two phases; first, the Sparse Factored Attention model in Section \ref{sec:sparse_architecture}
is trained using the discriminator task in Section \ref{sec:discriminator} for 200,000 steps with a learning rate of $10e^{-5}$
and batch size of 1024. Then, the weights at the end of training (for the corresponding initialization seed and $\mathcal{D}_{N}$
are frozen and used as the initialization for the VIN model described in Section \ref{sec:planning_vin}. The training parameters and setup
used otherwise is the same as is described in Appendix \ref{sec:appendix_baselines}.

\section{Additional Limitations}
\label{sec:appendix_limitations}

\paragraph{Controlled Environment} We used the \texttt{GoToLocal}
task on BabyAI as the sole reference environment for this study. A fully observable state space, knowledge of the
state-space connectivity, and disentangled factors on
cell states are very strong assumptions that are
leveraged to achieve the results that we present.

\paragraph{Computational resources}
Sample efficiency does not imply computational efficiency. In particular, we found that training
the discriminator in Section \ref{sec:discriminator} requires large
batch sizes and a large number of samples generated from $\mathcal{D}_{N}$
to converge. 

\section{Reproducibility of this work}
\label{sec:appendix_reproducibility}

We kept the importance of reproducible research in mind when designing our experimental
method. We provide the source code for our approach and seeds used to generate each
environment and trajectory in $\mathcal{D}$.

We are unable to provide pre-trained models or log files due to space constraints.

\section{Computational Resource usage of this work}
\label{sec:appendix_resource_usage}

The person responsible for developing the
method took about one year to do so and used a workstation with a single NVIDIA RTX2060 GPU
with 6GB of GPU memory to test different approaches. Because the methods
that we present in this paper may be sensitive to different weight initializations,
we believed it was necessary to show trained model performance using different
initialization random initializations, using the methods in \cite{journals/corr/abs-2108-13264/Agarwal/2021}
for a more reliable presentation of results. To conduct the experiments using the
final version of our methods, we used our SLURM compute cluster with an array
of shared NVIDIA Tesla V100 GPUs. We ran 6 different versions of the discriminator
experiment, over five different models, ten dataset sizes, ten random initializations, each one taking up
to 8 hours to complete, making for 24,000 hours of GPU time used. We ran 3 different versions
of the end-to-end experiments over 4 different models, with the same number of dataset sizes and random initializations
each one taking up to 12 hours, making for an additional 19,200 hours.

\end{document}